\pdfoutput=1
\documentclass[11pt]{article}

\usepackage{acl}
\usepackage[most]{tcolorbox}
\tcbset{
  promptstyle/.style={
    colback=gray!8,
    colframe=gray!60,
    coltitle=white,
    fonttitle=\bfseries,
    enhanced,
    boxrule=0pt,
    title={Prompt used for refinement:},
    arc=2pt,
    left=2mm,right=2mm,top=1mm,bottom=1mm
  }
}

\usepackage{latexsym}
\usepackage{url}
\usepackage{comment}
\usepackage{amsmath,amsthm,amsfonts,amssymb,bm,stmaryrd}
\usepackage{booktabs}
\usepackage{multicol, multirow}
\usepackage{algorithm}
\usepackage{colortbl}
\usepackage{algpseudocode}
\usepackage{algorithm}      
\usepackage{algpseudocode}  
\usepackage{tabularx, booktabs, colortbl} 
\usepackage{xcolor}
\usepackage{array}
\renewcommand{\arraystretch}{1.3}  
\usepackage{appendix}

\definecolor{rowgray}{gray}{0.94}

\usepackage{CJKutf8}
\usepackage{kotex}  

\usepackage[T1]{fontenc}

\usepackage[utf8]{inputenc}
\usepackage{booktabs, colortbl, siunitx, xcolor}
\usepackage{subcaption}
\usepackage{microtype}
\usepackage{graphicx}

\definecolor{color_table_feedback}{RGB}{253,242,208}

\title{RL from Teacher-Model Refinement: Gradual Imitation Learning for Machine Translation}

\renewcommand{\thefootnote}{\fnsymbol{footnote}}
\author{
  Dongyub Jude Lee \quad
  Zhenyi Ye \quad
  Pengcheng He\footnotemark[1]\footnotemark[2] \\
  Zoom Communications \\
  \texttt{\{jude.lee, zhenyi.ye, pengcheng.he\}@zoom.us}
}

\begin{document}
\urlstyle{sf}

\maketitle

\renewcommand{\thefootnote}{\arabic{footnote}}

\begin{abstract}
Preference-learning methods for machine translation (MT), such as Direct Preference Optimization (DPO), have shown strong gains but typically rely on large, carefully curated preference triplets and often struggle to generalize beyond their tuning domains. We propose \emph{Reinforcement Learning from Teacher-Model Refinement (RLfR)}, which replaces static triplets with \emph{on-policy, actor-conditioned refinements} produced by a frozen teacher. At each step, the actor samples candidate translations, the teacher performs a minimal local edit of each draft, and the actor is reinforced to close the gap using a composite reward that combines \emph{scaled negative edit distance} for lexical and structural fidelity with \emph{COMET} for semantic adequacy. This formulation yields a stable, model-aware learning signal without requiring explicit preference datasets. Experiments on FLORES-200 (\textit{en}$\leftrightarrow$\textit{de/es/zh/ko/ja}) show that RLfR consistently outperforms strong MT-SFT, DPO, and fixed-reference RL baselines, improving semantic quality and entity preservation.

\end{abstract}

\section{Introduction}
\label{sec:intro}
Recent breakthroughs in large language models (LLMs)~\cite{achiam2023gpt,grattafiori2024llama,team2024qwen2,yang2025qwen3} have markedly improved multilingual Machine Translation.
Models pre-trained on trillions of tokens in dozens of languages now
provide strong zero-shot and few-shot capabilities, pushing translation
quality far beyond traditional encoder–decoder systems and catalysing a
wave of research on how best to adapt these general-purpose LLMs to
production-grade MT~\cite{hoang2023fly,xu2023paradigm,alves2024tower,xu2024x}.

Machine Translation (MT) has advanced rapidly through
supervised fine-tuning (SFT) and, more recently,
preference-learning methods such as Direct Preference Optimization (DPO)~\cite{rafailov2023direct,agrawal2024modeling,tang2025mitigating,agrawal2024modeling,li2024mt,ramos2024fine}.
Although these approaches yield sizeable gains by leveraging explicit
feedback, they depend on large, carefully curated triplet
datasets—each built from an anchor translation and ranked alternatives.
Triplet construction is costly, slows iteration, and limits domain
generality.

Recent work exemplified by~\cite{donato2022mad,guo2025deepseek} explores off-policy reinforcement learning as an alternative to supervised fine-tuning, and has shown particular promise in structured domains such as mathematics and code generation. However, applying these methods to machine translation faces two critical obstacles. First, translation inherently lacks deterministic ground truth and admits multiple acceptable outputs for a given source sentence, making it difficult to formulate robust rule-based reward functions; lexical and syntactic diversity further reduces the utility of hand-crafted scoring functions. Second, off-policy updates that operate on static references or previously sampled responses often fail to deliver improvements commensurate with their computational cost, as these fixed targets can diverge substantially from the model’s predictions and in practice tend to yield convergence near the initial checkpoint’s performance (Table~\ref{tab:example_comparison}). Our proposed method addresses both challenges by delegating credit assignment to a frozen teacher model (e.g., GPT-4o-mini), which generates high-quality, context-sensitive refinements for each actor output. By conditioning the teacher’s feedback directly on the model’s own translations, we obtain minimal yet precise corrections that preserve the original sentence structure while progressively narrowing the quality gap between actor and teacher, thereby providing a more effective, model-aware reinforcement learning signal.

\begin{figure*}[t]
  \centering
  \includegraphics[width=\textwidth]{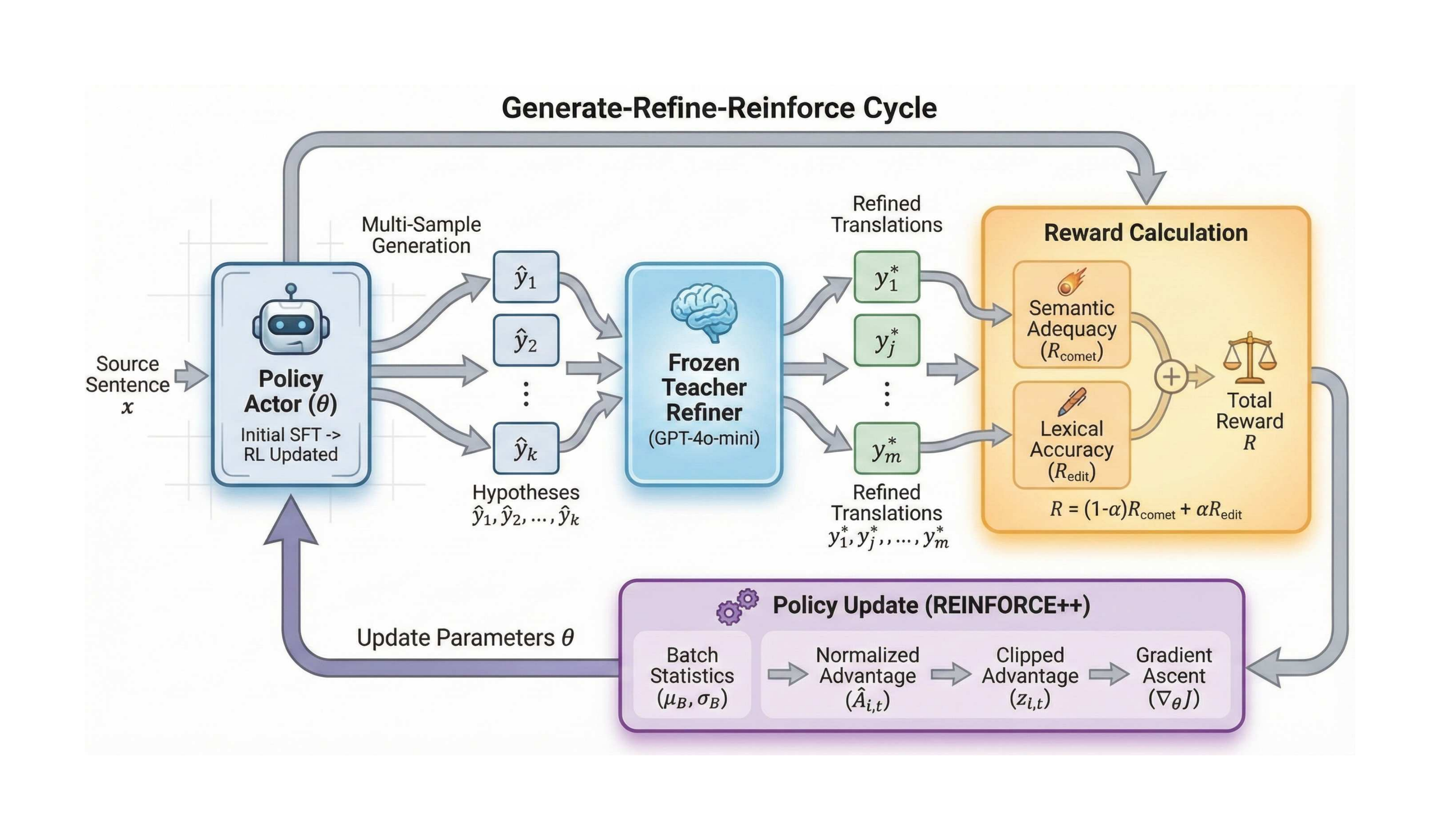}
  \caption{\textsc{RLfR} follows an incremental generate--refine--reinforce cycle. A policy actor is first initialised, then refined online by a frozen teacher model, and finally updated with a critic-free REINFORCE++ objective that uses multi-sampling.}
  \label{fig:rlfr_overview}
\end{figure*}

\smallskip
We present RL from Teacher-model Refinement (RLfR), a reinforcement-learning
framework that eliminates triplet construction.
For every source sentence, a frozen teacher model (GPT-4o-mini) refines the
actor’s draft translation \(\hat{y}\) into a higher-quality target
\(y^{\ast}\).
Lexical and semantic gaps between \(\hat{y}\) and \(y^{\ast}\) form the
\emph{learning signal} and are quantified by two complementary metrics:
(i) edit distance \(\bigl(R_{\mathrm{edit}}\bigr)\) and
(ii) COMET \(\bigl(R_{\mathrm{comet}}\bigr)\).
These scores are combined into a single reward, and a critic-free
REINFORCE++~\cite{hu2025reinforce++} update maximises the expected composite reward,
continually pushing the actor toward the refiner.
For evaluation only, we additionally report the M-ETA~\cite{conia-etal-2024-towards} metric to assess
entity-level fidelity. Our contribution can be summarized as follows:

\begin{itemize}
    \item \textit{Online refinement learning:}\,
          RLfR replaces static triplet data with real-time GPT-4o-mini
          refinements, removing dataset curation and frozen references.
    \item \textit{Progressive, multi-metric learning:}\,
          A single reward signal that blends edit-distance and COMET
          shrinks the actor–refiner gap at every step, delivering
          simultaneous gains in lexical fidelity, semantic adequacy,
          and entity correctness.
    \item \textit{Strong empirical gains:}\,
          On the FLORES-200 benchmark
          (en$\leftrightarrow$de/es/zh/ko/ja) RLfR improves COMET,
          reduces lexical errors, and lowers M-ETA entity mistakes,
          outperforming diverse baselines. 
\end{itemize}

These properties make RLfR a robust, scalable, and
data-efficient solution for multilingual MT. We will release the code and training scripts upon the acceptance of this paper.

\section{RL from Teacher-Model Refinement (\textsc{RLfR})}
\label{sec:rlfr}

\textsc{RLfR} follows an incremental generate--refine--reinforce cycle
(Fig.~\ref{fig:rlfr_overview}).
A supervised actor is first initialised, then repeatedly refined online by
a frozen teacher model, and finally updated with a critic-free
REINFORCE++ objective that uses multi-sampling.

\subsection{Supervised Actor Initialisation}
\label{sec:stage1}

Although pre-trained or initially supervised fine-tuned (SFT) multilingual models offer broad coverage, their performance on machine translation (MT) benchmarks frequently remains insufficient for high-quality evaluation settings. Prior studies~\cite{guo2025deepseek} further indicate that reinforcement learning (RL) benefits substantially from being initialized with a well-trained SFT model, rather than from a weaker or zero-shot baseline.

However, publicly available parallel corpora suffer from both data scarcity and inconsistent translation quality across language pairs. To obtain a stronger and more reliable starting point for RL, we leverage \textsc{GPT-4o-mini} to distill a high-quality multilingual parallel corpus. This synthetic supervision provides consistent quality while remaining significantly more cost-efficient than human-curated data or large-scale preference annotations (see Appendix~\ref{sec:teacher_cost_appendix}).

\begin{figure}[t]
  \centering
  \includegraphics[width=\linewidth]{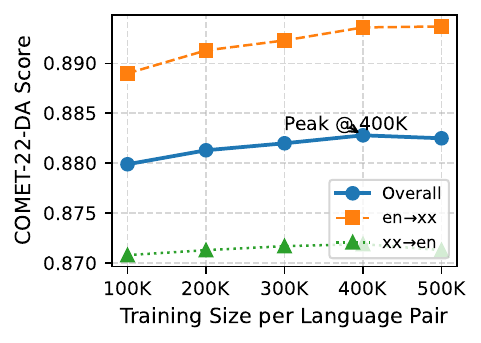}
  \caption{COMET scores on the FLORES test set using the LLaMA-3.1 8B model, evaluated with different amounts of distilled SFT training data per language pair.}
  \label{fig:sft_ablation}
\end{figure}

To assess how far static distillation alone can strengthen the base model, we fine-tune our actor using GPT-4o-mini–translated parallel datasets derived from the FineFineWeb corpus~\cite{finefineweb}. We construct identical samples across five language pairs and vary the amount of training data from 100K to 500K sentence pairs per language pair. This controlled setup allows us to directly measure the effect of increased offline teacher supervision on translation quality.

The actor parameters~$\theta$ are optimized from the initial checkpoints by minimizing the standard cross-entropy objective:
\begin{equation}
\mathcal{L}_{\text{SFT}} = \frac{1}{N} \sum_{i=1}^{B} \sum_{t=1}^{T_i} -\log P(y_{i,t} \mid x_i, y_{i,<t}),
\end{equation}
where $N$ denotes the number of predicting tokens in a mini batch.

As shown in Figure~\ref{fig:sft_ablation}, translation quality improves as more distilled data is used, but gains plateau beyond approximately 400K examples. Importantly, the 400K setting achieves the best overall COMET score (0.8828), whereas larger distillation sets yield no further gains despite more supervision budgets. While the marginal cost of synthetic pairs is low—and our online refinement is similarly inexpensive (Appendix~\ref{sec:teacher_cost_appendix})—the plateau suggests that the limitation of offline distillation is not cost but the static nature of its supervision. This, in turn, motivates more targeted, model-aware refinement signals as introduced in our RLfR framework.

Consequently, we adopt the 400K SFT model as the initialization for all subsequent RLfR experiments. In the following sections, we show that RLfR delivers substantial gains on top of this strong SFT baseline by incorporating \emph{online}, actor-conditioned refinements that cannot be captured through additional static distillation alone.

\subsection{Online Target Refinement via Multi-Sample Generation}
\label{sec:stage2}

Given a source sentence $x$, we first sample multiple translation hypotheses from the actor model:
\begin{equation}
  \hat{y}_{1:k} \sim p_\theta(y \mid x)
\end{equation}

Each hypothesis $\hat{y}_i$ is concatenated with the original source prompt $x$ and then passed to a frozen teacher refinement model $\mathcal{T}$ (GPT-4o-mini), yielding refined target translations:
\begin{equation}
  y_i^{\ast} = \mathcal{T}(x, \hat{y}_i)
\end{equation}

Unlike offline distillation, this refinement is \emph{actor-conditioned}: the teacher observes the model's hypothesis and produces a minimal, locally improved target rather than a static translation. This provides supervision that adapts to the actor's current behaviour and cannot be replicated by simply scaling distilled data.

By generating multiple hypotheses per input, the actor obtains a richer view of the local reward landscape, significantly reducing variance during policy gradient updates once batch-normalised advantages are computed.

\vspace{1ex}

\subsection{Reward Design and Policy Updates}
\label{sec:stage3}

\paragraph{Reward Formulation}
We design the reward function to balance semantic adequacy and lexical accuracy:

\begin{align}
R(\hat{y}, y^{\ast}) =\;& 
(1 - \alpha)\,\underbrace{R_{\mathrm{comet}}}_{\text{Semantic Adequacy}} \notag\\
& +\;\alpha\,\underbrace{R_{\mathrm{edit}}}_{\text{Lexical Accuracy}}, \quad \alpha \in [0,1]
\end{align}

Here, \( R_{\mathrm{edit}} \) is the scaled negative Levenshtein distance between the hypothesis \(\hat{y}\) and the refined target \(y^{\ast}\):
\begin{equation}
  R_{\mathrm{edit}} =
  s\left(1 - \frac{\operatorname{EditDist}(\hat{y}, y^{\ast})}{\max\{|\hat{y}|, |y^{\ast}|\}}\right),
\end{equation}
where the scaling function \( s(z) \) discretizes lexical similarity scores based on dynamic empirical quantiles calculated from a subset of the training corpus (\texttt{rollout\_batch\_size}):
\begin{equation}
  s(z)=
  \begin{cases}
    -1 & z < \text{mean}, \\[4pt]
    \frac{z - \text{mean}}{Q_{90} - \text{mean}} & \text{mean} \le z < Q_{90}, \\[4pt]
    1 & z \ge Q_{90},
  \end{cases}
\end{equation}

Here, \( \text{mean} \) and \( Q_{90} \) denote the average and the 90th percentile, respectively, dynamically computed from the subset samples.

The semantic reward \( R_{\mathrm{comet}} \) is the COMET~\cite{rei2022comet} score linearly mapped to \([-1,1]\). We explore three reward combinations: lexical-focused (\(\alpha=1\)), semantic-focused (\(\alpha=0\)), and balanced (\(\alpha=0.5\)).

\begin{algorithm}[t]
\caption{\textsc{RLfR}: Reinforcement Learning from Refinement}
\label{alg:rlfr}
\footnotesize
\begin{algorithmic}[1]
\Require 
    Mini-batch of inputs $\mathcal{B}$; 
    actor policy $\pi_\theta$ with parameters $\theta$; 
    frozen teacher-refiner $\mathcal{T}$; 
    number of samples $k$; 
    learning rate $\eta$
\Ensure Updated actor parameters $\theta$

\For{$x \in \mathcal{B}$} \Comment{Process each input in the batch}
    \State \textbf{Sample $k$ hypotheses} from the current actor:
        \[
            \hat{y}_{1:k} \sim \pi_\theta(\cdot \mid x)
        \]
    \State \textbf{Obtain refined targets} from the teacher:
        \[
            y^{\ast}_{1:k} \gets \mathcal{T}(x, \hat{y}_{1:k})
        \]
    \State \textbf{Compute per-sample rewards}:
        \[
            R_{1:k} \gets \textsc{ComputeReward}(\hat{y}_{1:k},\, y^{\ast}_{1:k})
        \]
\EndFor

\vspace{4pt}
\State \textbf{Compute batch-level statistics:}
    \[
        (\mu_B, \sigma_B) \gets \textsc{BatchStats}(R)
    \]

\State \textbf{Normalize advantages:}
    \[
        A_{1:k} \gets \frac{R_{1:k} - \mu_B}{\sigma_B}
    \]

\State \textbf{Update actor parameters using REINFORCE++:}
    \[
        \theta \gets \theta + \eta \,\nabla_\theta \mathcal{L}_{\text{R++}}(\theta; A_{1:k})
    \]

\end{algorithmic}
\end{algorithm}

\paragraph{Batch-normalised \textsc{REINFORCE++} Objective}

We now provide a detailed description of the RLfR training procedure, as summarized in Algorithm~\ref{alg:rlfr}. Let $B$ denote the minibatch of sampled hypotheses obtained from a single rollout, with each sample $i \in B$ associated with a scalar reward $R_i$. The advantage of each token $t$ in sequence $i$ of length $T_i$ is defined as:
\[
\begin{aligned}
A_{i,t}
&= R_i
- \beta \sum_{j=t}^{T_i}
\\
&\quad \mathrm{KL}\!\Bigl(
   \pi_\theta(\cdot|\hat{y}_{i,<j}, x_i)
   \,\|\, 
   \pi_{\theta_{\text{old}}}(\cdot|\hat{y}_{i,<j}, x_i)
\Bigr)
\end{aligned}
\]

where the Kullback-Leibler (KL) divergence term penalizes deviation from the reference policy $\pi_{\theta_{old}}$ for all tokens from position $t$ to the end of the sequence, and $\beta$ controls the strength of this regularization.

To stabilize the reward signal and mitigate variance during policy gradient estimation, we introduce a batch normalization strategy applied to the advantage. The batch-level advantage statistics are computed as follows:

\begin{equation}
\mu_B = \frac{1}{N}\sum_{i\in B}\sum_{t=1}^{T_i} A_{i,t},
\end{equation}

\begin{equation}
\sigma_B = \sqrt{\frac{1}{N}\sum_{i\in B}\sum_{t=1}^{T_i}(A_{i,t}-\mu_B)^2 + \varepsilon},
\end{equation}

where $\mu_B$ represents the mean advantage across the batch, $\sigma_B$ is the standard deviation, $N=\sum_{i\in B}T_i$ is the total number of tokens in the batch, and $\varepsilon$ is a small constant (e.g., $10^{-6}$) included for numerical stability.

Utilizing these batch-level statistics, we compute a batch-normalized advantage for each sample $i$ at position $t$:

\begin{equation}
\hat{A}_{i,t} = \frac{A_{i,t} - \mu_B}{\sigma_B}.
\end{equation}

This normalized advantage metric is subsequently employed to scale gradient updates on a per-sample basis, thereby stabilizing the learning process.

To prevent excessively large policy updates that might destabilize training, we further incorporate a clipped importance-weighted advantage inspired by Proximal Policy Optimization (PPO):

\begin{equation}
\begin{aligned}
\rho_{i,t} &= \frac{\pi_\theta(\hat{y}_{i,t}\mid x_i, \hat{y}_{i,<t})}
               {\pi_{\theta_{old}}(\hat{y}_{i,t}\mid x_i, \hat{y}_{i,<t})}, \\
z_{i,t}    &= \mathrm{clip}\bigl(\rho_{i,t},\,1-\epsilon,\,1+\epsilon\bigr)\,\hat{A}_{i,t}
\end{aligned}
\end{equation}

where $\pi_\theta(\hat{y}_{i,t}\mid x_i, \hat{y}_{i,<t})$ denotes the probability of selecting token $\hat{y}_{i,t}$ at position $t$ under the current policy parameterized by $\theta$, conditioned on the input $x_i$ and all previous tokens in the sequence, and $\pi_{\theta_{old}}(\hat{y}_{i,t}\mid x_i, \hat{y}_{i,<t})$ represents the corresponding probability under the reference policy—a snapshot of parameters prior to the most recent update. The clipping operation constrains the importance weight to the interval $[1 - \epsilon, 1 + \epsilon]$, thereby limiting the magnitude of individual policy updates at each token position and improving training stability.

Gradient ascent is then performed using the batch-normalized and clipped policy gradient estimation across all tokens in the batch, defined as:

\begin{equation}
\nabla_{\theta}J(\theta) = \frac{1}{N}\sum_{i\in B}\sum_{t=1}^{T_i} z_{i,t} \cdot \nabla_{\theta}\log\pi_{\theta}(\hat{y}_{i,t}\mid x_i,\hat{y}_{i,<t}).
\end{equation}

This formulation of the objective significantly enhances training stability and convergence properties, effectively balancing exploration and exploitation while ensuring robust policy updates.

\begin{table*}[t]
  \centering
  \small
  \resizebox{\textwidth}{!}{%
  \begin{tabular}{
    l
    *{5}{S[table-format=2.3]}
    S[table-format=2.2, detect-weight=true, mode=text]
    *{5}{S[table-format=2.3]}
    S[table-format=2.3, detect-weight=true, mode=text]
  }
    \toprule
    & \multicolumn{6}{c}{\textbf{COMET (\%) — XX$\to$EN}} & \multicolumn{6}{c}{\textbf{COMET (\%) — EN$\to$XX}} \\
    \cmidrule(lr){2-7} \cmidrule(lr){8-13}
    \textbf{Baseline \& Method} & {de} & {es} & {ja} & {ko} & {zh} & \textbf{Avg.}
                                & {de} & {es} & {ja} & {ko} & {zh} & \textbf{Avg.} \\
    \midrule
    \multicolumn{13}{l}{\emph{LLaMA-3.1-8B as Baseline}} \\
    \quad Baseline & 88.72 & 86.29 & 87.18 & 87.28 & 86.46 & 87.19
                   & 88.73 & 87.10 & 91.73 & 90.37 & 88.85 & 89.36 \\
    \rowcolor{gray!10}
    \quad \textbf{Semantic-Focused} & 89.64 & 87.55 & 88.44 & 88.36 & 87.60 & \bfseries 88.32
                                    & 88.87 & 87.41 & 91.90 & 90.54 & 89.00 & \bfseries 89.54 \\
    \quad Lexical-Focused & 89.50 & 87.34 & 88.12 & 88.19 & 87.27 & 88.08
                       & 88.79 & 87.34 & 91.74 & 90.38 & 88.89 & 89.43 \\
    \quad Mix & 89.55 & 87.47 & 88.28 & 88.29 & 87.51 & 88.23
              & 88.82 & 87.38 & 91.81 & 90.46 & 88.80 & 89.45 \\
    \midrule
    \multicolumn{13}{l}{\emph{Qwen3-1.7B as Baseline}} \\
    \quad Baseline & 88.50 & 86.01 & 87.07 & 86.76 & 86.40 & 86.95
                   & 87.71 & 86.41 & 91.02 & 89.68 & 88.53 & 88.67 \\
    \rowcolor{gray!10}
    \quad \textbf{Semantic-Focused} & 88.96 & 86.49 & 87.62 & 87.46 & 86.98 & 87.50
                                    & 88.04 & 86.86 & 91.30 & 90.18 & 88.73 & \bfseries 89.02 \\
    \quad Lexical-Focused & 88.85 & 86.38 & 87.62 & 87.45 & 86.94 & 87.45
                       & 87.95 & 86.62 & 91.26 & 90.05 & 88.58 & 88.89 \\
    \quad Mix & 89.02 & 86.47 & 87.69 & 87.58 & 87.07 & \bfseries87.57
              & 87.92 & 86.66 & 91.24 & 90.06 & 88.69 & 88.91 \\

    \midrule
    \multicolumn{13}{l}{\emph{ZLM-2.3B as Baseline}} \\
    \quad Baseline & 88.36 & 86.02 & 86.86 & 87.01 & 85.97 & 86.84
                   & 88.06 & 86.53 & 91.48 & 89.97 & 88.08 & 88.82 \\
    \rowcolor{gray!10}
    \quad \textbf{Semantic-Focused} & 89.11 & 86.95 & 87.69 & 87.96 & 86.78 & \bfseries 87.70
                                    & 88.03 & 86.88 & 91.72 & 90.27 & 88.48 & \bfseries 89.08 \\
    \quad Lexical-Focused & 88.98 & 86.79 & 87.57 & 87.74 & 86.62 & 87.54
                       & 88.06 & 86.74 & 91.57 & 90.01 & 88.29 & 88.93 \\
    \quad Mix & 89.09 & 86.93 & 87.68 & 88.00 & 86.77 & 87.69
              & 88.17 & 86.81 & 91.63 & 90.11 & 88.38 & 89.02 \\
    \bottomrule
  \end{tabular}
  }
  \caption{COMET (\%) scores for both XX$\to$EN and EN$\to$XX directions across three SFT baselines. Bolded values indicate best average scores in each block.}
  \label{tab:comet_combined}
\end{table*}

\section{Experiments}

\begin{table*}[t]
\centering
\scalebox{0.90}{%
\small
\begin{tabular}{
    l
    S[table-format=2.2]
    S[table-format=2.2]
    S[table-format=2.2]
    S[table-format=2.2]
    S[table-format=1.2]
    S[table-format=2.2]
    c
}
    \toprule
    & \multicolumn{6}{c}{\textbf{M-ETA (\%) per Language}} & \\
    \cmidrule(lr){2-7}
    \textbf{Baseline \& Method} & {de} & {es} & {ja} & {ko} & {zh} & {\textbf{Avg.}} & {\textbf{Gain}} \\
    \midrule
    \multicolumn{8}{l}{\emph{LLaMA-3.1-8B as Baseline}} \\
    \quad Baseline & 28.45 & 40.60 & 26.42 & 32.89 & 4.43 & 26.56 & \multicolumn{1}{c}{---} \\
    \rowcolor{gray!10}
    \quad \textbf{Mix} & 33.65 & 45.47 & 28.08 & 37.05 & 4.29 & \bfseries 29.71 & \textbf{(+3.15)} \\
    \quad Lexical-Focused & 33.11 & 46.68 & 27.11 & 36.78 & 4.43 & 29.62 & (+3.06) \\
    \quad Semantic-Focused & 32.01 & 44.93 & 27.52 & 37.58 & 4.72 & 29.35 & (+2.79) \\
    \midrule
    \multicolumn{8}{l}{\emph{Qwen3-1.7B as Baseline}} \\
    \quad Baseline & 23.94 & 33.83 & 21.02 & 25.91 & 4.85 & 21.91 & \multicolumn{1}{c}{---} \\
    \rowcolor{gray!10}
    \quad \textbf{Mix} & 24.49 & 33.69 & 22.41 & 29.4 & 4.99 & \bfseries 23.00 & \textbf{(+1.09)} \\
    \quad Lexical-Focused & 24.35 & 35.18 & 22.41 & 28.05 & 4.57 & 22.91 & (+1.00) \\
    \quad Semantic-Focused & 23.53 & 33.83 & 22.54 & 26.31 & 4.71 & 22.18 & (+0.27) \\
    \midrule
    \multicolumn{8}{l}{\emph{ZLM-2.3B as Baseline}} \\
    \quad Baseline & 21.20 & 35.45 & 20.47 & 27.25 & 3.32 & 21.54 & \multicolumn{1}{c}{---} \\
    \rowcolor{gray!10}
    \quad \textbf{Mix} & 22.57 & 38.02 & 23.65 & 31.81 & 3.74 & \bfseries 23.96 & \textbf{(+2.42)} \\
    \quad Lexical-Focused & 21.61 & 36.94 & 23.24 & 31.95 & 3.74 & 23.50 & (+1.96) \\
    \quad Semantic-Focused & 22.85 & 37.21 & 21.16 & 31.68 & 4.02 & 23.38 & (+1.84) \\
    \bottomrule
\end{tabular}
} 
\caption{M-ETA (\%) performance across five languages under three SFT baselines. Each block reports baseline and enhanced methods. Best-performing variants are bolded.}
\label{tab:meta_baseline_comparison}
\end{table*}

\subsection{Experimental Setup}
\label{sec:exp_setup}

\textbf{Supervised MT-SFT Baselines.}  
We fine-tune several pre-trained LLMs---such as LLaMA-3.1-8B~\cite{grattafiori2024llama}, Qwen3-1.7B~\cite{yang2025qwen3}, and ZLM-2.3B~\cite{zoom2025agentic}---on distilled parallel data comprising 400K sentence pairs for each translation direction (\( \text{en} \rightarrow \text{xx} \), \( \text{xx} \rightarrow \text{en} \)). Each model is optimized via cross-entropy loss to learn a strong supervised policy \( p_\theta(y \mid x) \).

\textbf{Off-line Preference Learning.}  
Starting from each MT-SFT baseline, we generate two candidate translations for every source prompt: one produced by the MT-SFT model and one supplied by GPT-4o-mini as a distilled reference. We then apply reference-free COMET to score each \(\langle\text{prompt},\text{candidate}\rangle\) pair. The candidate with the higher COMET score is labeled as “positive” and the other as “negative.” These static triplets are subsequently used to train the model via Direct Preference Optimization (DPO), isolating the gains achieved purely through preference learning.

\textbf{RL with Fixed References.}  
From each MT-SFT checkpoint, we sample 5K examples per direction from the same distilled corpus and treat the GPT-4o-mini outputs as immutable ground truth. Actors are updated using the standard REINFORCE++ objective against these fixed rewards to quantify the benefits of RL when the teacher’s outputs remain static.

\textbf{RL from Refinement (RLfR, proposed).}  
Using each MT-SFT checkpoint, we subsample 5K prompts per direction and draw \(k\) hypotheses from the actor. Each hypothesis is sent to teacher for refinement, and actors are trained with batch-normalised REINFORCE++ loss using dynamic rewards. This configuration evaluates the efficacy of continuous, high-quality feedback from a teacher external model.

All experiments—supervised, off-line preference learning, and reinforcement learning—use subsets of the same distilled dataset to ensure a controlled and directly comparable evaluation. Teacher outputs are generated using GPT-4o-mini with a fixed low-temperature decoding setup; full configuration details are provided in Appendix~\ref{sec:teacher_config}.

\subsection{Experimental Results}
\label{sec:results}

\textbf{RLfR is Effective in Improving Semantic Precision and Fluency.}
\label{sec:results_semantic_precision}
In this section, we evaluate the semantic precision improvements achieved through our RLfR method. Our experiments encompassed two main categories based on model size: a large model (LLaMA-3.1-8B) and smaller models (Qwen3-1.7B and ZLM-2.3B). Semantic precision was measured using COMET scores, reflecting nuanced semantic adequacy and fluency across multiple translation directions (XX\textrightarrow EN and EN\textrightarrow XX).

As summarized in Table~\ref{tab:comet_combined}, our RLfR method consistently enhances semantic performance for both large and smaller baseline models. For the large model (LLaMA-3.1-8B), the Semantic-Focused RLfR approach achieves the highest average COMET scores in both directions, surpassing Lexical-Focused and Mixed variants. This notable improvement highlights RLfR's capacity to reinforce semantic coherence and translation fluency.

Similarly, for the smaller models, RLfR demonstrates consistent gains across both translation directions. The Semantic-Focused variant proves particularly effective, offering steady improvements over the baselines and reinforcing the generalizability of our method across model sizes.

\begin{figure}[t]
\centering
\small

\begin{tabularx}{\linewidth}{>{\bfseries}lX}
\rowcolor{rowgray}
Prompt & How does Jia Yingchun contribute to the overall narrative of the Dream of the Red Chamber? \\
\end{tabularx}

\vspace{0.6em}

\begin{tabularx}{\linewidth}{>{\bfseries}p{3cm}X}
\toprule
Reference &
\begin{CJK}{UTF8}{gbsn}贾迎春对红楼梦的整体叙述有何贡献？\end{CJK} \newline
\begin{CJK}{UTF8}{gbsn}贾\end{CJK} (Jia: family name), \begin{CJK}{UTF8}{gbsn}迎\end{CJK} (welcome), \begin{CJK}{UTF8}{gbsn}春\end{CJK} (spring) \\
\midrule
SFT Output &
\begin{CJK}{UTF8}{gbsn}贾莹春如何为《红楼梦》的整体叙述做出贡献？\end{CJK} \newline
\begin{CJK}{UTF8}{gbsn}莹\end{CJK} (crystal or glittering): semantically incorrect \\
\midrule
RLfR Output &
\begin{CJK}{UTF8}{gbsn}贾英春如何为《红楼梦》的整体叙述做出贡献？\end{CJK} \newline
\begin{CJK}{UTF8}{gbsn}英\end{CJK} (heroic or outstanding): phonetically accurate, semantically appropriate \\
\bottomrule
\end{tabularx}

\caption{Comparison of translations for “Jia Yingchun.” SFT introduces semantic distortion, whereas RLfR produces a phonetically faithful and contextually valid variant.}
\label{fig:rlfr_entity_case}
\end{figure}

\noindent\textbf{RLfR Effectively Preserves Named Entities Across Languages.}\label{sec:results_entity_preservation} Accurate translation of named entities is critical in maintaining factual consistency and user trust, especially in high-stakes applications such as legal, medical, and technical domains. Even minor errors in entity preservation can lead to significant misunderstandings. To address this challenge, we evaluate the effectiveness of our RLfR approach in improving entity preservation using the M-ETA metric. The evaluation covers both a large model and smaller models, across five target languages.

As shown in Table~\ref{tab:meta_baseline_comparison}, RLfR consistently enhances M-ETA scores across all model scales. Improvements are particularly pronounced in the large model, where all configurations show strong gains over the baseline. Among them, the mixed configuration—combining semantic-focused and lexical-focused objectives—achieves the highest overall performance, indicating that jointly optimizing both aspects leads to better entity preservation.

For the smaller models, RLfR also yields consistent improvements, though the gains are relatively smaller. The mixed strategy again performs best, followed by either the semantic-focused or lexical-focused variants depending on the model. These results suggest that while RLfR benefits models of all sizes, larger models tend to realize greater improvements.


While RLfR yields substantial M-ETA gains across most target languages, we observe relatively smaller improvements in Chinese. This discrepancy is not due to RLfR being less effective for Chinese, but rather reflects the limitations of reference-based automatic metrics. In our dataset, many Chinese references—especially for named entities—are inconsistently rendered or localized. For example, as described in Figure ~\ref{fig:rlfr_entity_case}, the name "Jia Yingchun" is translated in the reference as "\begin{CJK}{UTF8}{gbsn}贾迎春\end{CJK}" (\textit{\begin{CJK}{UTF8}{gbsn}贾\end{CJK}}: Jia [family name], \textit{\begin{CJK}{UTF8}{gbsn}迎\end{CJK}}: welcome, \textit{\begin{CJK}{UTF8}{gbsn}春\end{CJK}}: spring), which is the canonical form found in classical literature. However, RLfR produces "\begin{CJK}{UTF8}{gbsn}贾英春\end{CJK}" (\textit{\begin{CJK}{UTF8}{gbsn}贾\end{CJK}}: Jia [family name], \textit{\begin{CJK}{UTF8}{gbsn}英\end{CJK}}: heroic or outstanding, \textit{\begin{CJK}{UTF8}{gbsn}春\end{CJK}}: spring), which is phonetically faithful and semantically appropriate in modern usage. In contrast, SFT outputs "\begin{CJK}{UTF8}{gbsn}贾莹春\end{CJK}" (\textit{\begin{CJK}{UTF8}{gbsn}莹\end{CJK}}: crystal or lustrous), where the middle character introduces a semantic shift that alters the intended identity. Despite being arguably more natural and appropriate than the reference in a modern context, RLfR's output is penalized by M-ETA for deviating from the gold-standard reference. This case illustrates how rigid references and character-level overlap metrics may underestimate real-world gains in entity preservation—particularly in languages like Chinese where orthographic variation is common.

Overall, our findings highlight the effectiveness of RLfR in preserving named entities during translation, especially when both semantic and lexical-based signals are combined in training.

\section{Analysis}

\subsection{Effectiveness over Baseline Approaches}

We compare our proposed RLfR against two baselines: DPO and RL with Fixed References. All results are based on an ablation study using the LLaMA-3.1-8B model.

\begin{table}[ht]
\centering
\footnotesize
\renewcommand{\arraystretch}{1.15}
\resizebox{0.49\textwidth}{!}{%
\begin{tabular}{lcc}
\toprule
\textbf{Method} & \textbf{COMET (Avg.)}  & \textbf{M-ETA (Avg.)}  \\
\midrule
MT-SFT (baseline) & 88.23 & 26.56 \\
DPO & 88.63 & 27.19 \\
RL w/ Fixed Ref. & 88.80 & 28.20 \\
\rowcolor{gray!12}
\textbf{RLfR (Ours)} & \textbf{88.93} & \textbf{29.35} \\
\bottomrule
\end{tabular}}
\vspace{3pt}
\caption{Comparison of average COMET and M-ETA scores using LLaMA-3.1-8B. Best-performing results are bolded.}
\label{tab:comparison_methods}
\end{table}

Table~\ref{tab:comparison_methods} summarizes the aggregated scores and the findings can be summarized as:

\begin{itemize}
\item \textbf{RLfR achieves the strongest overall performance, with the highest COMET and M-ETA.} This suggests that refinement using dynamic feedback from a stronger model enhances both fluency and entity preservation.

\item \textbf{RL with Fixed References performs moderately well.} It surpasses DPO but still lags behind RLfR, likely because fixed references rely on fully rewritten teacher outputs that are often misaligned with the model’s own predictions. As the reward encourages similarity to a static target rather than improvements over the model’s actual output, the learning signal becomes less relevant and harder to optimize. Additional qualitative examples are provided in Appendix Table~\ref{tab:example_comparison}.

\item \textbf{DPO yields the lowest scores,} highlighting its limitations in the absence of explicit correction signals. Preference-based comparisons without an improved target offer limited guidance for fine-grained control.
\end{itemize}

These findings highlight RLfR’s effectiveness in improving semantic adequacy and entity fidelity through iterative, model-aware feedback from a teacher model.

\subsection{LLM-as-a-Judge Evaluation}
\label{sec:llm_judge}

To complement automatic metrics, we perform an LLM-as-a-judge evaluation
over 500 examples (250 En$\rightarrow$X and 250 X$\rightarrow$En across
five languages). We use \texttt{gpt-5-mini} as the judge model. Importantly,
\emph{this judge model is completely independent from the teacher used
during RLfR training}, which relies solely on \texttt{gpt-4o-mini} for
all refinement steps. This separation ensures that evaluation is not
biased toward the teacher model and that improvements are assessed by a
distinct, unbiased evaluator.

The judge scores four criteria—semantic adequacy, entity fidelity,
fluency, and hallucination (higher indicates fewer unsupported additions)—
and also provides an overall pairwise preference. As shown in
Tables~\ref{tab:llmjudge_metrics} and~\ref{tab:llmjudge_preference},
RLfR achieves the strongest results across all evaluation dimensions.
It delivers better meaning preservation, more accurate entity handling,
higher fluency, and substantially fewer hallucinations than SFT, DPO,
and Fixed-Reference RL. Pairwise preferences consistently favor RLfR,
reinforcing the gains observed in automatic metrics.

\begin{table}[t]
\centering
\small
\renewcommand{\arraystretch}{1.05}
\setlength{\tabcolsep}{3pt}
\resizebox{\columnwidth}{!}{%
\begin{tabular}{lcccc}
\toprule
\textbf{Method} &
\textbf{Sem.} &
\textbf{Entity} &
\textbf{Fluency} &
\textbf{Halluc.~(↑)} \\
\midrule
SFT              & 4.358 & 4.476 & 4.300 & 4.632 \\
DPO              & 4.534 & 4.572 & 4.375 & 4.705 \\
RL w/ Fixed Ref. & 4.500 & 4.580 & 4.460 & 4.740 \\
\rowcolor{gray!12}
RLfR (Ours)      & \textbf{4.570} & \textbf{4.636} &
                    \textbf{4.548} & \textbf{4.836} \\
\bottomrule
\end{tabular}}
\caption{
Fine-grained LLM-as-a-judge scores over 500 examples.
RLfR achieves the highest performance across all criteria, with
Fixed-Reference RL falling between DPO and RLfR.
Scores were assigned by \texttt{gpt-5-mini}, which is independent from
the \texttt{gpt-4o-mini} teacher used in RLfR training.
}
\label{tab:llmjudge_metrics}
\end{table}

\begin{table}[t]
\centering
\small
\renewcommand{\arraystretch}{1.05}
\resizebox{\columnwidth}{!}{%
\begin{tabular}{lccc}
\toprule
\textbf{Comparison} &
\textbf{RLfR Win} &
\textbf{Tie} &
\textbf{Baseline Win} \\
\midrule
RLfR vs.\ SFT        & 231 & 126 & 143 \\
RLfR vs.\ DPO        & 200 & 121 & 179 \\
RLfR vs.\ Fixed Ref. & 215 & 140 & 145 \\
\bottomrule
\end{tabular}}
\caption{
Pairwise preferences from 500 judged comparisons per baseline.
RLfR is consistently preferred over all alternatives.
Judgments were produced by \texttt{gpt-5-mini}, ensuring evaluation by
an independent model unrelated to the \texttt{gpt-4o-mini} teacher.
}
\label{tab:llmjudge_preference}
\end{table}

\subsection{Qualitative Comparison of Supervision Methods}

To demonstrate the difference in supervision dynamics, Table~\ref{tab:example_comparison} shows an example comparing a model's prediction, a fixed teacher reference, and an RLfR-refined version. All outputs are generated from the same Korean input describing a Netflix series.

This example highlights a key limitation of fixed references: they represent complete rewrites from the teacher model, often diverging significantly from the model's original phrasing and structure. Although these references are grammatically sound and semantically correct, their dissimilarity limits their utility as reward targets, since they fail to indicate what the model needs to improve. In contrast, RLfR conditions the teacher model on the model's actual output and query, enabling targeted and minimal refinements. This produces revised responses that retain the model’s structure but correct factual and stylistic errors—thereby offering more effective, model-aware guidance for reinforcement learning.

\subsection{Stepwise Analysis of Teacher Mimicry}
\label{sec:teacher_mimicry}

\begin{figure*}[t]
  \centering
  \begin{subfigure}[t]{0.32\textwidth}
    \centering
    \includegraphics[width=\linewidth]{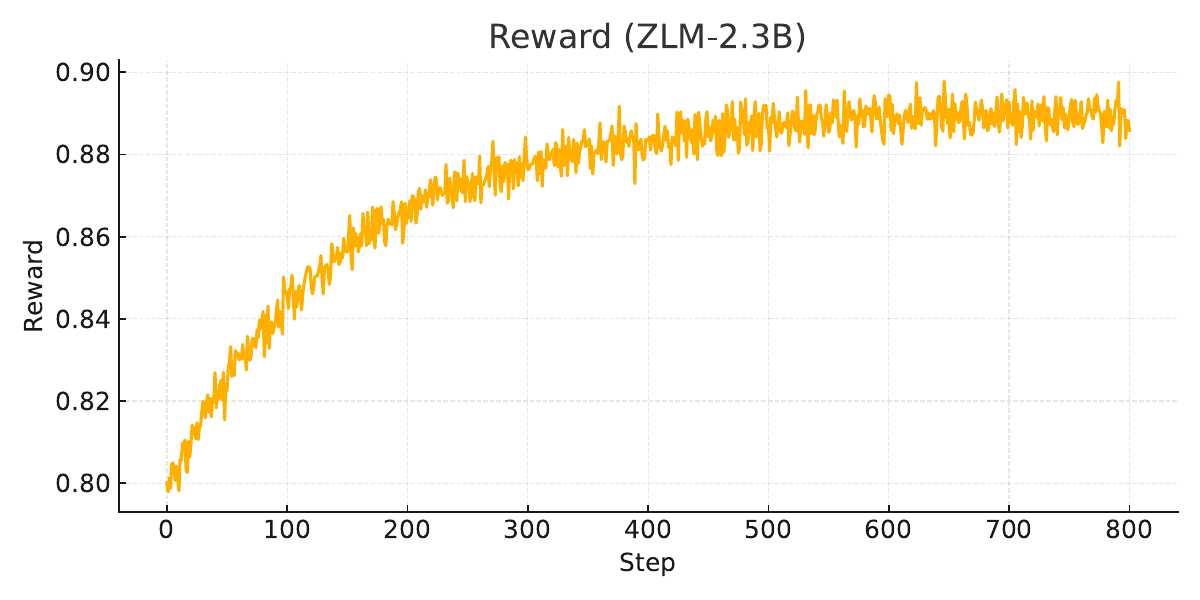}
    \caption{Reward trajectory}
    \label{fig:sub_reward}
  \end{subfigure}\hfill
  \begin{subfigure}[t]{0.32\textwidth}
    \centering
    \includegraphics[width=\linewidth]{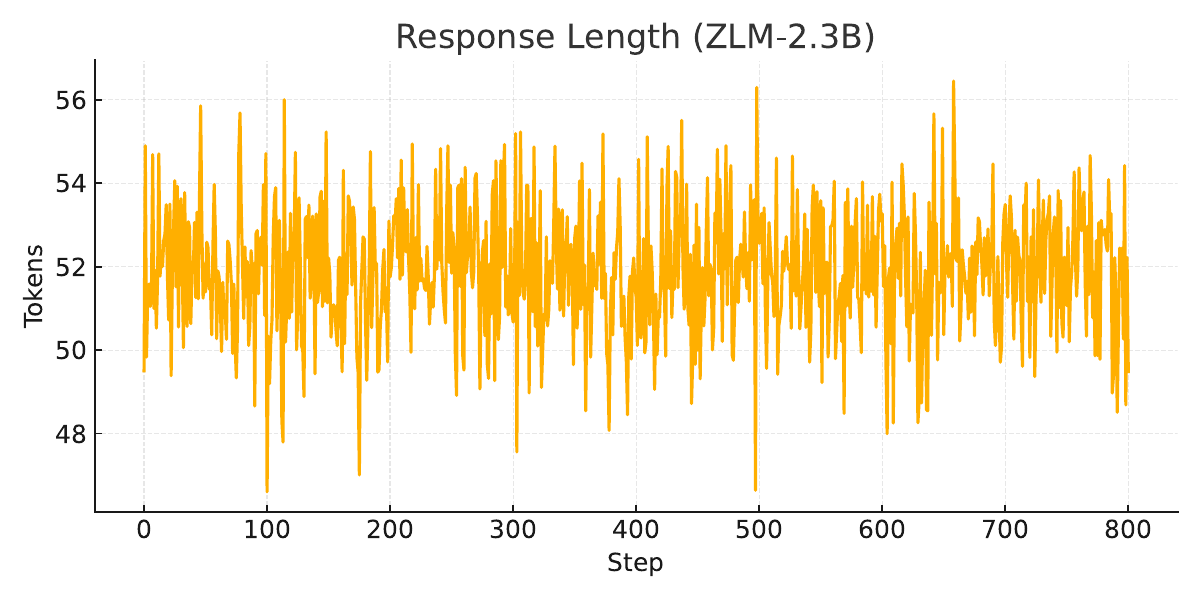}
    \caption{Response-length stability}
    \label{fig:sub_resp_len}
  \end{subfigure}\hfill
  \begin{subfigure}[t]{0.32\textwidth}
    \centering
    \includegraphics[width=\linewidth]{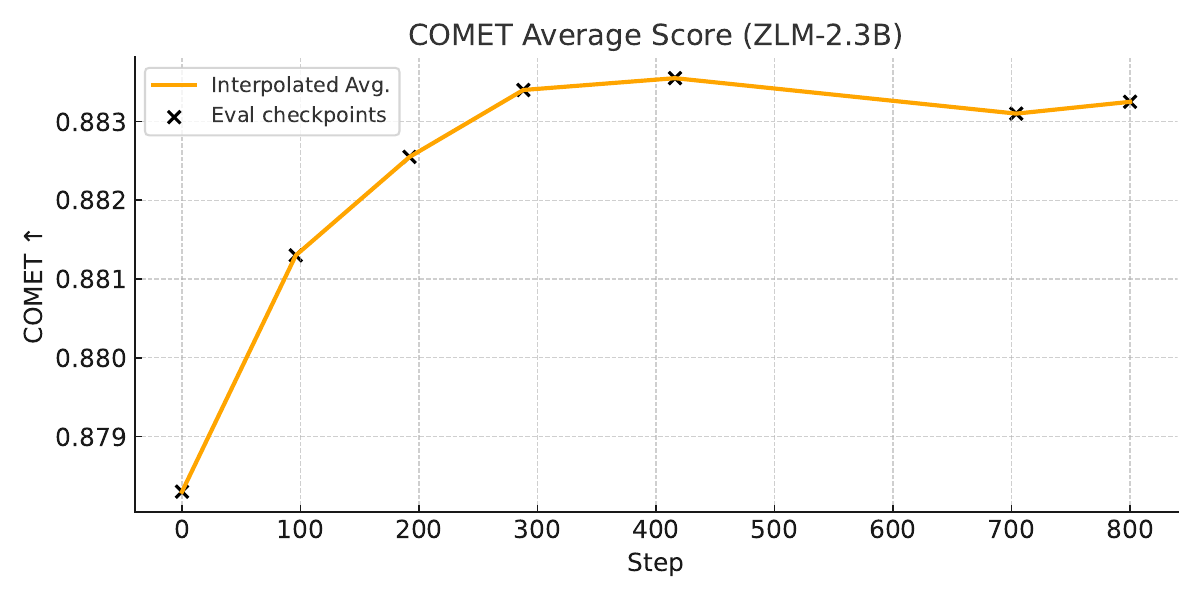}
    \caption{COMET-22 (avg)}
    \label{fig:sub_comet}
  \end{subfigure}
  \caption{Stepwise dynamics of teacher-guided refinement for the ZLM-2.3B model trained with a mixture of supervised and refinement-based updates. Reward rises monotonically, response length stays within a narrow band, and COMET-22 peaks at step~416—aligning with the reward plateau—before marginally tapering off.}
  \label{fig:teacher_mimicry_triple}
\end{figure*}

To better understand how our model internalizes the teacher's refinements, we analyze three core training signals in Figure~\ref{fig:teacher_mimicry_triple}: the teacher-assigned reward, the response-only length, and the COMET score.

As shown in Figure~\ref{fig:sub_reward}, the reward steadily increases throughout training, indicating that the model progressively refines its responses to better align with the teacher’s preferred outputs. This consistent upward trend, without signs of early saturation, demonstrates that the actor effectively internalizes the teacher’s corrective signal and improves through structured, stepwise mimicry.

Figure~\ref{fig:sub_resp_len} demonstrates that the model does not exploit token length to inflate reward. Response lengths remain tightly bounded throughout training, which suggests that the reward gains stem from substantive improvements rather than degenerate strategies such as output expansion or pruning. The model appears to preserve its original structure while focusing on localized refinements.

As shown in Figure~\ref{fig:sub_comet}, the COMET average score steadily increases throughout training and peaks around step 400, consistent with the teacher reward trend. This improvement reflects that the model not only mimics the teacher’s edits effectively but also delivers outputs with stronger semantic adequacy and lexical precision. Notably, the score remains stable even beyond the peak, showing no sign of reward overfitting or regression—indicating that improvements are durable and generalizable. This validates our stepwise refinement objective as a stable and effective supervision strategy.

\subsection{Qualitative Case Study: Successes and Failures}
\label{sec:entity_semantic_case_multilingual}

While automatic metrics such as COMET and M-ETA quantify overall performance, they may overlook nuanced improvements in specific linguistic dimensions. To complement these metrics, we present multilingual qualitative examples that demonstrate both the strengths and limitations of RLfR compared to the SFT baseline. We highlight two aspects: (1) entity-level preservation, where RLfR better maintains named entities, and (2) semantic-level adequacy, where RLfR improves fluency and contextual meaning. An illustrative example for Korean is provided in Table~\ref{tab:case_kr}. Additionally, we include representative failure cases to foster transparency and guide future improvements.

\section{Related Work}

\textbf{LLM-based Machine Translation}  
The success of large language models (LLMs) in various NLP tasks has led to increased use in machine translation. Recent studies have primarily focused on enhancing multilingual capabilities and translation quality. Approaches include augmenting multilingual capacities~\cite{yang2023bigtranslate, cui2025multilingual}, integrating reasoning capabilities~\cite{liu2025new}, and improving translation quality through fine-tuning and knowledge distillation~\cite{xu2023paradigm, li2024mt, stap2024fine}. Additionally, methods employing plug-and-play modules, adaptive rejection mechanisms~\cite{xu2024x}, and gradient-based data selection~\cite{pan2024g} have been explored. Research has also investigated scaling laws~\cite{isik2025scaling} and advancements in document-level translation leveraging large-scale corpora~\cite{pal2024document}.

\textbf{Preference Learning in Machine Translation}  
Beyond supervised fine-tuning, preference-based learning incorporates human feedback or fine-grained criteria to improve translation outputs. PPO~\cite{schulman2017proximal} has been used with quality estimation models~\cite{he2024improving}, and span-level rewards from xCOMET~\cite{guerreiro2024xcomet} enable token-level fine-grained reward modeling~\cite{ramos2024fine}. DPO~\cite{rafailov2023direct} has been extended with PPO~\cite{zhong2024dpo}, and automatic metrics (e.g., XCOMET) are used to build preference datasets~\cite{agrawal2024modeling}. ~\citet{xu2024contrastive} proposed Contrastive Preference Optimization (CPO), while also using COMET-based ranking for reward selection\cite{he2024improving}. ~\citet{li2024mt} leverages LLM-generated feedback and analogy data for targeted distillation. However, most of these approaches rely on well-curated triplets or fixed references, which may produce outputs disconnected from the model’s current predictions.

\section{Conclusion}
\label{sec:conclusion}

We presented \textsc{RLfR}, a reinforcement learning framework that replaces static preference triplets with continuous, high-quality refinements from a teacher model. By rewarding both negative edit distance and COMET, \textsc{RLfR} jointly encourages lexical fidelity and semantic adequacy while naturally annealing reward magnitudes as the actor improves. 

Across five FLORES-200 language pairs, \textsc{RLfR} consistently surpasses strong MT-SFT, DPO, and fixed-reference RL baselines, yielding substantial gains in COMET and M-ETA across models of diverse scales. These results demonstrate that iterative teacher feedback is a more sample-efficient and robust alternative to static preference data for MT. 

Future work includes extending \textsc{RLfR} to broader language families and integrating richer feedback signals such as discourse coherence and domain adaptation cues.

\section*{Limitations}
\paragraph{Teacher Bias}
Because RLfR depends on teacher-generated translations, any biases or systematic errors in the teacher model can be directly inherited by the student. This makes the strength and calibration of the teacher particularly important for preventing the amplification of undesirable behaviors.

\paragraph{Teacher Invocation Cost}
Although RLfR requires multiple teacher queries during refinement, the actual cost depends on how the teacher is accessed. When using API-based models, monetary cost can be a concern, whereas locally deployable strong models reduce API cost and allow batching or hardware-level optimizations. Nonetheless, large-scale refinement still demands careful resource–performance trade-offs.

\bibliography{anthology,custom}
\bibliographystyle{acl_natbib}

\appendix




\clearpage
\begin{appendices}
\label{sec:appendix}

\section{Implementation Details}

\textbf{Teacher Generation Settings}  
\label{sec:teacher_config}
All teacher outputs used for data distillation, preference learning, and
refinement are generated using \texttt{gpt-4o-mini} under a fixed
low-temperature, deterministic-like decoding setup (temperature = 0.1,
top-p = 1.0, max\_tokens = 2048). We use the default chat-completions
API interface, and all prompts follow the system–user format shown below
to ensure consistency across all stages of training.

\begin{tcolorbox}[promptstyle]
You are an expert in Machine Translation and Cultural Localization. Refine a given machine-translated sentence by improving its quality and naturally localizing named entities into the target language only.

\textbf{Guidelines}
\begin{itemize}
  \item \textbf{Entity Localization}: Translate or adapt all named entities into natural, target-language-only forms. Do not include the source language. Use official or culturally standard localized names.
  \item \textbf{Style and Tone}: Match the tone of the source (formal, neutral, conversational).
  \item \textbf{Fluency}: Ensure the result is natural, idiomatic, and fluid.
  \item \textbf{Grammar and Accuracy}: Fix any awkward grammar or inconsistent wording.
\end{itemize}

\end{tcolorbox}

\vspace{0.5em}

\section{Teacher Cost for \texttt{gpt-4o-mini}}
\label{sec:teacher_cost_appendix}

We estimate the monetary cost of the refinement stage, which is the only
component of RLfR requiring repeated teacher queries. On average, each
teacher input contains 63 tokens when averaged across the five target
languages (de, es, ja, ko, zh) and both
$\text{en}\rightarrow\text{xx}$ and $\text{xx}\rightarrow\text{en}$
directions. The RL dataset consists of 5K prompts per direction, but the
actual number of \texttt{gpt-4o-mini} calls is determined by the rollout
configuration. At each RL step we sample
\texttt{rollout\_batch\_size} $=560$ prompts and draw
$n_{\text{samples}} = 8$ hypotheses per prompt, all of which are refined
by the teacher. For $S$ RL steps this yields
\[
N_{\text{calls}} = S \times 560 \times 8,
\]
which corresponds to $N_{\text{calls}} \approx 1{,}792{,}000$ teacher
invocations when $S = 400$.

Assuming the \texttt{gpt-4o-mini} pricing scheme with per-million prices
$p_{\text{in}} = \$0.15$ for input tokens and $p_{\text{out}} = \$0.60$
for output tokens, we define the per-token costs
$c_{\text{in}} = p_{\text{in}} / 10^6$ and
$c_{\text{out}} = p_{\text{out}} / 10^6$.
Denoting the average refinement length by $L_{\text{out}}$, the total
cost of the refinement stage is
\[
\text{Cost}_{\text{refine}}
= N_{\text{calls}} \bigl(63\, c_{\text{in}} + L_{\text{out}}\, c_{\text{out}}\bigr).
\]
For translation, the output length is typically comparable to the input
length across the five language pairs, so we approximate
$L_{\text{out}} \approx 63$. This yields roughly $1.1\times 10^8$ input
tokens and $1.1\times 10^8$ output tokens, corresponding to a total cost
of about \$80--90 for a single RLfR run with $S = 400$ steps. The same
formula applies to any alternative teacher, with only the pricing
parameters changing.

\section{Alternative Teacher Models and Plug-and-Play Refinement}
\label{sec:alt_teacher_appendix}

RLfR treats the teacher as a black-box refiner: the algorithm only
assumes access to a model that can improve or rewrite candidate
translations, and does not depend on a specific architecture or family.
To assess robustness to the teacher choice, we run small-scale
experiments with alternative refiners for the LLaMA-3.1-8B baseline,
including full \texttt{gpt-4o} and \texttt{gpt-5} in addition to our
default \texttt{gpt-4o-mini} teacher.

To keep teacher cost manageable, we restrict these ablations to the
Korean and Chinese directions. For each teacher, we fine-tune RLfR with
the same hyperparameters and report COMET scores obtained by averaging
XX$\rightarrow$EN and EN$\rightarrow$XX for each language (ko, zh),
which we use as our main COMET metrics in this appendix.

\begin{table}[t]
\centering
\small
\begin{tabular}{lc}
\toprule
Teacher &
\begin{tabular}{@{}c@{}}
COMET $\uparrow$ \\
ko $\mid$ zh
\end{tabular} \\
\midrule
SFT Baseline (no RLfR)      & 88.83 $\mid$ 87.66 \\
RLfR (\texttt{gpt-4o-mini}) & 89.38 $\mid$ 88.16 \\
RLfR (\texttt{gpt-4o})      & 89.39 $\mid$ 88.16 \\
RLfR (\texttt{gpt-5})       & 89.42 $\mid$ 88.18 \\
\bottomrule
\end{tabular}
\caption{
COMET scores for RLfR with alternative teacher models on LLaMA-3.1-8B.
We evaluate only Korean (ko) and Chinese (zh) to control teacher cost.
For each language, we report a single COMET score obtained by averaging
the XX$\rightarrow$EN and EN$\rightarrow$XX directions; each cell shows
ko $\mid$ zh.
All RLfR settings improve over the SFT baseline; \texttt{gpt-4o-mini}
and \texttt{gpt-4o} yield almost identical performance, while the
higher-capacity \texttt{gpt-5} provides larger gains.
}
\label{tab:alt_teacher_llama}
\end{table}

All three teachers produce very similar reward trajectories and stable
RL updates throughout training. Consistent with
Table~\ref{tab:alt_teacher_llama}, \texttt{gpt-4o-mini} and
\texttt{gpt-4o} are nearly indistinguishable in COMET, and
\texttt{gpt-5} achieves only slightly higher scores at a higher
inference cost. In practice, \texttt{gpt-4o-mini} therefore offers a
substantially more cost-efficient option while exhibiting only marginal
degradation in refinement quality on these benchmarks.

We note that \texttt{gpt-5} is currently our strongest general-purpose
teacher model. However, this work was finalized in late May 2025,
before the public release of \texttt{gpt-5}. As a result, we did not
re-run all main experiments with \texttt{gpt-5} as the default refiner
and instead include it only in this small-scale ablation. A more
comprehensive comparison with \texttt{gpt-5} teachers is left for
future work.

Overall, these results indicate that RLfR is \emph{teacher-agnostic}:
it can flexibly adopt higher-capacity, more economical, or
domain-specialized refinement models, as long as they provide reasonably
strong machine translation quality. This plug-and-play design makes it
straightforward to swap in future teacher models without changing the
core RLfR algorithm.

\section{Qualitative Analysis: Successes and Failures}

This appendix expands upon the multilingual qualitative case study 
presented in Section~\ref{sec:entity_semantic_case_multilingual}. 
In addition to the core examples discussed in the main text, we include 
a broader set of success cases—such as improved entity preservation and 
semantic-level adequacy—as well as illustrative failure cases that reveal 
remaining challenges for RLfR. These extended examples provide a more 
comprehensive view of where RLfR improves over the SFT baseline and where 
it still struggles, thereby complementing automatic metrics and supporting 
transparency.

\begin{table}[ht]
\centering
\footnotesize
\renewcommand{\arraystretch}{1.3}
\resizebox{0.49\textwidth}{!}{%
\begin{tabular}{p{0.95\linewidth}}
\toprule
\textbf{Input (Korean)} \\
넷플릭스 시리즈 '오징어 게임' 시즌3의 제작발표회가 9일 오전 서울 용산구 드래곤시티에서 열렸다.\\ 
시즌3에서는 다시 게임에 참여하게 된 기훈, 신분을 숨기고 침투하는 프런트맨, 그리고 살아남은 참가자들의 마지막 운명이 그려진다. \\
\midrule

\textbf{SFT Model Output (Draft)} \\
The third season of the Netflix series \textcolor{red}{'Ongiye Game'} was officially launched at Seoul Dragon City on September 9 in the morning. 
The \textcolor{red}{Ozone Game} Season 3 tells the story of \textcolor{red}{'Gintetsu'}, a game participant who loses his closest friend after rejoining the game with a new purpose, 'Frontman', who hides his identity and enters the game, and the final fate of the participants who survive in this brutal game. \\
\midrule

\textbf{RLfR-Refined Output} \\
The third season of the Netflix series \textcolor{blue}{Squid Game} was officially unveiled at Seoul Dragon City on the morning of September 9. 
The new season follows \textcolor{blue}{Gi-hun}, who reenters the \textcolor{blue}{deadly game} with a renewed purpose but ends up losing his closest friend; 
the mysterious \textcolor{blue}{Front Man}, who conceals his identity while infiltrating the game; 
and the final fate of the remaining participants who survive its brutal trials. \\
\midrule

\textbf{Fixed Reference from Teacher} \\
On the morning of the 9th, a production presentation for Netflix’s Squid Game Season 3 was held at Seoul Dragon City in Yongsan-gu, Seoul. 
Squid Game Season 3 follows the story of Gi-hun, who reenters the game with a personal mission but ends up losing his closest friend; 
the Front Man, who hides his identity while infiltrating the game; 
and the final destinies of the survivors of the brutal competition. \\
\bottomrule
\end{tabular}}
\vspace{3pt}
\caption{
Comparison of outputs under different supervision styles. 
The \textcolor{red}{SFT Model Output} contains name errors and awkward phrases (e.g., \textcolor{red}{“Ongiye Game”}, \textcolor{red}{“Gintetsu”}, \textcolor{red}{“Ozone Game”}).
The \textcolor{blue}{RLfR-refined output} corrects these into \textcolor{blue}{“Squid Game”}, \textcolor{blue}{“Gi-hun”}, and improves fluency (e.g., \textcolor{blue}{“deadly game”}, \textcolor{blue}{“conceals his identity”}) while preserving the structure of the original response—providing a more effective and model-aware reward signal.
}

\label{tab:example_comparison}
\end{table}

\begin{table*}[t]
  \centering
  \renewcommand{\arraystretch}{1.25}
  \begin{tabular}{@{}p{0.48\textwidth}@{\hspace{0.02\textwidth}}p{0.48\textwidth}@{}}
    \toprule
    \textbf{Entity‑Level Success (en$\rightarrow$kr)} & \textbf{Semantic‑Level Success (kr$\rightarrow$en)} \\
    \midrule
    \textbf{Source:} In its decision, the PUB said it had concerns about bill affordability for Manitoba consumers, recommending the province development a program to assist lower-income Hydro customers. &
    \textbf{Source:} 젊은 음악가였던 시절, 토마스는 이 지역에서 잘 알려진 로이 헤드와 트레잇츠의 로이 헤드와 친구가 되어 함께 활동했습니다. \\

    \textbf{SFT:} 판결에서 PUB는 매니토바 소비자들의 청구서 affordability에 우려를 표명하며, 저소득 하이드로 고객을 지원하는 프로그램을 개발할 것을 권고했습니다. &
    \textbf{SFT:} While he was a young musician, Thomas befriended Roy Head of Roy Head and the Traits, who was well known in the area. \\

    \textbf{RLfR:} PUB는 판결에서 \textbf{매니토바 소비자들의 \underline{요금 부담에 대한 우려를 표명}}하며, \textbf{저소득층 하이드로 고객을 지원하는 프로그램을 개발할 것}을 권고했습니다. &
    \textbf{RLfR:} During his time as a young musician, Thomas \textbf{became friends with Roy Head from the well-known local band, Roy Head and the Traits}, and they \textbf{worked together}. \\

    \textcolor{blue}{Comment: RLfR \underline{correctly rephrases} \texttt{"bill affordability"} as \texttt{"요금 부담"} and \underline{preserves} named entities like PUB, Manitoba, and Hydro — showing stronger fidelity.} &
    \textcolor{blue}{Comment: RLfR \underline{retains and refines} the relationship and sequence of events, adding clarity while preserving meaning.} \\
    \midrule

    \textbf{Entity‑Level Failure (en$\rightarrow$kr)} & \textbf{Semantic‑Level Failure (kr$\rightarrow$en)} \\
    \midrule
    \textbf{Source:} 12x12" oil on canvas panel. &
    \textbf{Source:} 구리. \\

\textbf{RLfR:} 30x30\,cm \textbf{캔버스 패널에 유화.} &
    \textbf{RLfR:} \textbf{Guri.} \\

    \textcolor{red}{Comment: RLfR \underline{converts imperial to metric} (\texttt{"12x12 in"}$\rightarrow$\texttt{"30x30 cm"}), \underline{altering source meaning} — a fidelity loss.} &
    \textcolor{red}{Comment: RLfR \underline{mistranslates} the chemical element \texttt{copper} as the Korean city \textbf{“Guri”} based on phonetics — a clear semantic error.} \\
    \bottomrule
  \end{tabular}
  \caption{Korean (\texttt{kr}) examples showcasing RLfR’s strengths (blue) and weaknesses (red) in entity fidelity and semantic accuracy.}
  \label{tab:case_kr}
\end{table*}

\end{appendices}

\end{document}